# An Algorithm for the Construction of Bayesian Network Structures from Data

**Moninder Singh**
Computer Science Department
University of South Carolina
Columbia, SC 29208
< msingh@usceast.cs.scarolina.edu >

**Marco Valtorta**
Computer Science Department
University of South Carolina
Columbia, SC 29208
< mgv@usceast.cs.scarolina.edu >

## Abstract

Previous algorithms for the construction of Bayesian belief network structures from data have been either highly dependent on conditional independence (CI) tests, or have required an ordering on the nodes to be supplied by the user. We present an algorithm that integrates these two approaches - CI tests are used to generate an ordering on the nodes from the database which is then used to recover the underlying Bayesian network structure using a non CI based method. Results of preliminary evaluation of the algorithm on two networks (ALARM and LED) are presented. We also discuss some algorithm performance issues and open problems.

## 1   INTRODUCTION

In very general terms, different methods of learning probabilistic network structures from data can be classified into three groups. Some of these methods are based on linearity and normality assumptions ([Glymour et. al., 87], [Pearl & Wermuth, 93]); others are more general but require extensive testing of independence relations ([Fung & Crawford, 90], [Verma & Pearl, 92], [Spirtes & Glymour, 91], [Pearl & Verma, 91], [Spirtes, Glymour & Scheines, 90]); others yet take a Bayesian approach ([Herskovits, 91], [Cooper & Herskovits, 92], [Lauritzen, Thiesson & Spiegelhalter, 93]).

In this paper, we do not consider methods of the first kind, namely, those that make linearity and normality assumptions. Our work concentrates on CI test based methods and Bayesian methods. A number of algorithms have been designed which are based on CI tests. However, there are two major drawbacks of such CI test based algorithms. Firstly, the CI test requires determining independence relations of order $n - 2$, in the worst case. "Such tests may be unreliable, unless the volume of data is enormous" [Cooper & Herskovits, 92, page 332]. Also, as Verma and Pearl [Verma & Pearl, 92, pages 326-327] have noted, "in general, the set of all independence statements which hold for a given domain will grow exponentially as the number of variables grow". As such, CI test based approaches become rapidly computationally infeasible as the number of vertices increases. [Spirtes & Glymour, 91, page 62] have presented "an asymptotically correct algorithm whose complexity for fixed graph connectivity increases polynomially in the number of vertices, and may in practice recover sparse graphs with several hundred variables"; but for dense graphs with limited data, the algorithm might be unreliable [Cooper & Herskovits, 92].

On the other hand, [Cooper & Herskovits, 92] have given a Bayesian non-CI test based method, which they call the BLN (Bayesian learning of belief networks) method. Given that a set of four assumptions hold ([Cooper & Herskovits, 92, page 338]), namely, (i) The database variables are discrete, (ii) Cases occur independently, given a belief network model, (iii) All variables are instantiated to some value in every case, and finally (iv) Before observing the database, we are indifferent regarding the numerical probabilities to place on the belief network structure, Cooper and Herskovits have shown the following result:

**Theorem 1.**
[Cooper & Herskovits, 92]. Let $Z$ be a set of $n$ discrete variables, where variable $x_i$ in $Z$ has $r_i$ possible value assignments: $(v_{i1}, \ldots, v_{ir_i})$. Let $D$ be a database of $m$ cases, where each case contains a value assignment for each variable in $Z$. Let $B_S$ denote a belief network structure containing just the variables in $Z$. Each variable $x_i$ in $B_S$ has a set of parents $\pi_i$. $w_{ij}$ denotes the $j$th unique instantiation of $\pi_i$ relative to D and there are $q_i$ such unique instantiations of $\pi_i$. $N_{ijk}$ is the number of cases in D in which $x_i$ has the value $v_{ik}$ while $\pi_i$ is instantiated to $w_{ij}$. Let $N_{ij} = \sum_{k=1}^{r_i} N_{ijk}$. Then,

$$P(B_S, D) = P(B_S) \prod_{i=1}^{n} g(i, \pi_i) \qquad (1)$$

where $g(i, \pi_i)$ is given by

$$g(i, \pi_i) = \prod_{j=1}^{q_i} \frac{(r_i - 1)!}{(N_{ij} + r_i - 1)!} \prod_{k=1}^{r_i} N_{ijk}! \quad (2)$$

□

This result can be used to find the most probable network structure given a database. However, since the number of possible structures grow exponentially as a function of the number of variables, it is computationally infeasible to find the most probable belief network structure, given the data, by exhaustively enumerating all possible belief network structures.

Herskovits and Cooper ( [Cooper & Herskovits, 92], [Herskovits, 91] ) proposed a greedy algorithm, called the **K2** algorithm, to maximize $P(B_S, D)$ by finding the parent set of each variable that maximizes the function $g(i, \pi_i)$. In addition to the four assumptions stated above, K2 uses two more assumptions, namely, that there is an ordering available on the variables and that, a priori, all structures are equally likely. The K2 algorithm considers each node in the order given to it as input and determines its parents as follows. It first assumes that a node has no parents, and then adds incrementally that node (among the predecessors in the ordering) as a parent which increases the probability of the resultant structure by the largest amount. It stops adding parents to the node when the addition of no additional single parent can increase the probability.

## 2    MOTIVATION

As stated at the end of the previous section, the K2 algorithm requires an ordering on the nodes to be given to it as an input along with the database of cases. The main thrust of this research is to combine both CI as well as non CI test based methods described above to come up with a computationally tractable algorithm which is not overdependent on the CI tests, nor does it require a node ordering[1].

In order to achieve this, we use CI tests to generate an ordering on the nodes, and then use the K2 algorithm to generate the underlying belief network from the database of cases using this ordering of nodes. Also, since we are interested in recovering the most probable Bayesian network structure given the data, we would like to generate an ordering on the nodes that is consistent with the partial order specified by the nodes of the underlying network. In a domain where very little expertise is available, or the number of vertices is fairly large, finding such an ordering may not be feasible. As such, we would like to avoid such a requirement. The remainder of this section elaborates on this point.

It is possible to find a Bayesian network for any given ordering of the nodes, since any joint probability distribution $P(x_1, x_2, \ldots, x_n)$ can be rewritten, by successive applications of the chain rule, as $P(x_{i1}, x_{i2}, \ldots, x_{in}) = P(x_{i1} \mid x_{i2}, \ldots, x_{in}) \times P(x_{i2} \mid x_{i3}, \ldots, x_{in}) \times \ldots \times P(x_{in})$, where $< i_1, i_2, \ldots, i_n >$ is an arbitrary permutation of $< 1, 2, \ldots, n >$. However, the sparseness of the Bayesian network structure representing the joint probability distribution $P(x_1, x_2, \ldots, x_n)$ will vary, sometimes dramatically, with respect to the choice of the ordering of the nodes[2]. It is desirable to use an ordering of the nodes that allows as many of the conditional independences true in the probability distribution describing the domain of interest as possible to be represented graphically[3].

It would be too expensive to search blindly among all orderings of nodes, looking for one that leads to a network that both fits the data and is sparse enough to be useful. In a small setting, grouping variables into generic classes, such as symptoms and diseases may be sufficient to limit the number of orderings to be searched without having to use dramatically greedy heuristics. This was shown to be adequate for a medical application with 10 nodes in [Lauritzen, Thiesson, and Spiegelhalter, 1993], where variables were divided in "blocks." In some applications, however, it may be impossible to divide variables into classes, or the classes may be too large to impose sufficient structure on the space of candidate orderings. We have implemented an algorithm, called **CB**,[4] that uses a CI test based algorithm to propose a total order of the nodes that is then used by a Bayesian algorithm. We have tested the algorithm on some distributions generated from known Bayesian networks. (The results will be shown after the algorithm is presented.)

The Bayesian method used in the CB algorithm is a slightly modified version of Cooper and Herskovits's K2, implemented in C on a DECstation 5000. Herskovits proved an important result concerning the correctness of the metric that K2 uses to guide its search. He showed that the metric on which K2 is based is minimized, as the number of cases increases, without limit, on "those [Bayesian] network structures that, *for a given node order*, most parsimoniously capture all the independencies manifested in the data" [Herskovits, 1991, chapter 6]. More precisely, he showed that the K2 metric will always favor, as the number of cases in the database increase without limit, a minimal

---

[1]Herskovits [Herskovits, 91] suggested the use of the metric (on which K2 is based) with a CI test based method to do away with the requirement for an order of nodes.

[2]In this paper, no distinction is made between the nodes of a Bayesian network and the variables they represent.

[3]Whereas different types of graphical structures have different expressive powers, this paper is only concerned with dags, as used in Bayesian nets. We ignore Markov nets [Pearl, 88, chapter 3], chain graphs [Lauritzen and Wermuth, 1989a; 1989b], and other graphical representations (e.g., [Shachter, 1991; Geiger and Heckerman, 1991]).

[4]The name reflects the initials of the two phases of the algorithm.

I-map consistent with the given ordering (see [Pearl, 1988, chapter 3] for the definition of minimal I-map). Despite the convergence result, it is still important to provide K2 with a good node order, since there are too many orderings (n! for n nodes) to search blindly among them, unless drastically greedy (myopic) search regimens are used. Moreover, for different orderings, we will get different I-maps of differing density. Note that an I-map only means that all independencies implied by it (through d-separation) are also in the underlying model. So more sparse networks will give us more information as compared to relatively denser networks. In this sense, the ordering given to K2 becomes very important. Given a random ordering, we might land up with a very dense dag which is an I-map (possibly minimal) but conveys very little information. So, we would like to use as informed an ordering as possible. For example, assuming that the data was generated using a Bayesian network whose structure is an I-map for the underlying distribution, it would be very desirable to provide K2 with an ordering of the nodes that allows the network to be recovered exactly, even though K2 may recover a different I-map when given a different ordering, because the generating structure is normally the sparsest one among all I-maps for a given distribution, or at least one of the sparsest ones. Our algorithm finds good node orderings by using a CI-based test. Since CB still uses K2 to compute the Bayesian network structure from an ordering, it is correct in the same sense that K2 is.

# 3  DISCUSSION OF THE ALGORITHM

## 3.1  OVERVIEW

The algorithm basically consists of two phases: Phase I uses CI tests to generate an undirected graph, and then orients the edges to get an ordering on the nodes. Phase II takes as input a total ordering consistent with the DAG generated by phase I, and applies the K2 algorithm to construct the network structure using the database of cases. The two phases are executed iteratively - first for 0th order CI relations, then for 1st order CI relations, and so on until the termination criteria is met.

Steps 1 to 4 of the algorithm are based on the algorithms given by ([Verma & Pearl, 92] and [Spirtes & Glymour, 91]). We have allowed edges to be oriented in both directions because at any given stage, since CI tests of all orders have not been performed, all CI relations have not been discovered and there will be a number of extra edges. In such a case, it is quite possible for edges to be oriented in both directions by step 3. Although the bound used in step 2 is not necessary, it may be useful in decreasing the run time of the algorithm by not trying to generate the belief network structure if the undirected graph recovered from very low order CI relations (in step 2) is dense.

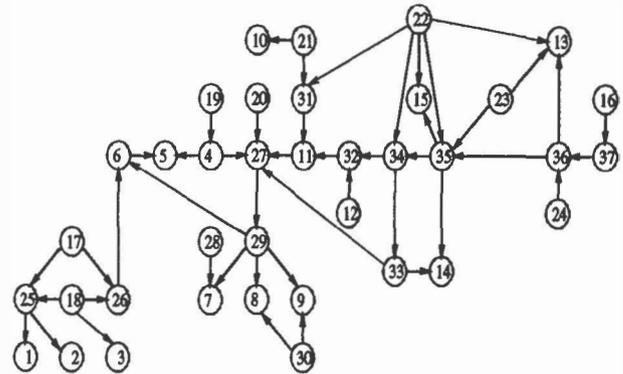

Figure 1: The ALARM Network

Once the edges have been oriented by steps 3 and 4, the algorithm finds the set of potential parents of each node by considering only the directed edges (step 5), and then uses a heuristic to choose an orientation for the edges which are still undirected, or are bidirected. Although, theoretically, equation 1 can be used to find the probability $P(\ i\ \rightarrow\ j\ |\ D)$ (and $P(\ i\ \leftarrow\ j\ |\ D)$) from the data ([Cooper & Herskovits, 92, page 318]) which can then be used to orient an edge $i - j$ (on the basis of which orientation is more probable), it is computationally infeasible do so because of the sheer number of network structures which have that edge. Hence the use of a heuristic. From equation 1, it should be clear that the orientation of an edge between vertices $i$ and $j$ affects only $g(i, \pi_i)$ and $g(j, \pi_j)$, and so to maximize the product $g(i, \pi_i) \times g(j, \pi_j)$ where $\pi_i$ and $\pi_j$ are the sets of parents of nodes $i$ and $j$ respectively. Accordingly, we compute the products $i_{val} = g(i, \pi_i) \times g(j, \pi_j \cup \{i\})$ and $j_{val} = g(j, \pi_j) \times g(i, \pi_i \cup \{j\})$ where $\pi_i$ and $\pi_j$ are the sets of potential parents recovered by step 5 of the algorithm. These products give us a way of selecting an orientation for the edge. If $i_{val}$ is larger, we prefer the edge $i\ \rightarrow\ j$ (unless it causes a directed cycle in which case we choose the other orientation). Similarly, we choose $j\ \rightarrow\ i$ if $j_{val}$ is larger (or the reverse in case of a directed cycle).

At this stage, the algorithm has constructed a DAG. It then finds a total ordering on the nodes consistent with the DAG and applies the K2 algorithm to find the set of parents of each node such that the K2 metric (i.e. $g(i, \pi_i)$) is maximized for each node $i$, allowing edges to be directed from a node only to nodes that are its successors in the ordering.

## 3.2  THE ALGORITHM

Let $A_G ab$ be the set of vertices adjacent to $a$ or $b$ in graph $G$ not including $a$ and $b$. Also, let $u$ be a bound on the degree of the undirected graph generated by step 2. *ord* is the order of CI relations being tested. Let $\pi_i$ be the set of parents of node $i, 1 \leq i \leq n$.

1. Start with the complete graph $G_1$ on the set of vertices $Z$.

   $ord \leftarrow 0$.

   $old\_\pi_i \leftarrow \{\ \}\ \forall i, 1 \leq i \leq n$, and $old\_Prob \leftarrow 0$.

2. [Spirtes & Glymour, 91]

   Modify $G_1$ as follows :

   For each pair of vertices $a, b$ that are adjacent in $G_1$, if $A_{G_1}ab$ has a cardinality greater than or equal to *ord*, and $I(a, S_{ab}, b)$ [5] where $S_{ab} \subseteq A_{G_1}ab$ of cardinality *ord*, remove the edge $a - b$, and store $S_{ab}$.

   If for all pairs of adjacent vertices $a, b$ in $G_1$, $A_{G_1}ab$ has cardinality $< ord$, goto step 10.

   If degree of $G_1 > u$, then

   $ord \leftarrow ord + 1$

   Goto beginning of step 2

3. Let $G$ be a copy of $G_1$.

   For each pair of non adjacent variables $a, b$ in $G$, if there is a node $c$ that is not in $S_{ab}$ and is adjacent to both $a$ and $b$, then orient the edges from $a \rightarrow c$ and $b \rightarrow c$ ([Verma & Pearl, 92], [Spirtes & Glymour, 91]) unless such an orientation leads to the introduction of a directed cycle in the graph.

   If an edge has already been oriented in the reverse direction, make that edge bidirected.

4. Try to assign directions to the yet undirected edges in $G$ by applying the following four rules [Verma & Pearl, 92], if this can be done without introducing directed cycles in the graph:

   Rule 1: If $a \rightarrow b$ and $b - c$ and $a$ and $c$ are not adjacent, then direct $b \rightarrow c$.

   Rule 2: If $a \rightarrow b$, $b \rightarrow c$ and $a - c$, then direct $a \rightarrow c$.

   Rule 3: If $a - b$, $b - c$, $b - d$, $a \rightarrow d$, and $c \rightarrow d$, then direct $b \rightarrow d$.

   Rule 4: If $a - b$, $b - c$, $a - c$, $c - d$, and $d \rightarrow a$, then direct $a \rightarrow b$ and $c \rightarrow b$.

   Moreover, if $a \rightarrow b, b \rightarrow c$ and $a \leftrightarrow c$, then direct $a \rightarrow c$.

5. Let $\pi_i \leftarrow \{\ \}\ \forall i, 1 \leq i \leq n$.

   For each node $i$, add to $\pi_i$ the set of vertices $j$ such that for each such $j$, there is an edge $j \rightarrow i$ in the pdag $G$.

6. For each undirected or bidirected edge in the pdag $G$ choose an orientation as described below

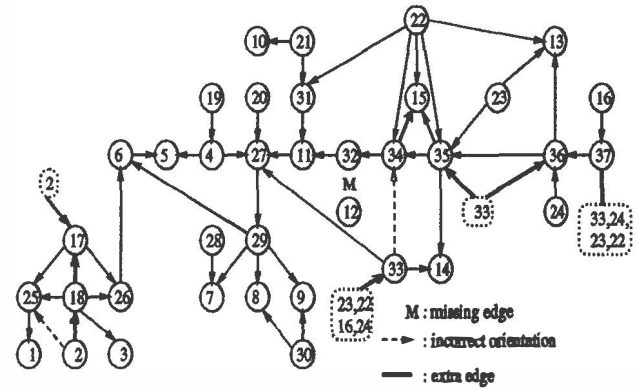

Figure 2: Constructed ALARM (Total)

If $i - j$ in an undirected edge, and $\pi_i$ and $\pi_j$ are the corresponding parent sets in $G$, then calculate the following products

$i_{val} = g(i, \pi_i) \times g(j, \pi_j \cup \{i\})$

$j_{val} = g(j, \pi_j) \times g(i, \pi_i \cup \{j\})$

If $i_{val} > j_{val}$, then $\pi_j \leftarrow \pi_j \cup \{i\}$ unless the addition of this edge, i.e. $i \rightarrow j$ leads to a cycle in the pdag. In that case, choose the reverse orientation, and change $\pi_i$ (instead of $\pi_j$). Do a similar thing in case $j_{val} > i_{val}$

7. The sets $\pi_i$, $1 \leq i \leq n$ obtained by step 6 define a DAG since for each node $i$, $\pi_i$ consists of those nodes that have a directed edge to node $i$.

   Generate a total order on the nodes from this DAG by performing a topological sort on it.

8. Apply the K2 algorithm to find the set of parents of each node using the order in step 7. Let $\pi_i$ be the set of parents, found by K2, of node $i$, $\forall 1 \leq i \leq n$.

   Let $new\_Prob = \prod_{i=1}^{n} g(i, \pi_i)$.

9. If $new\_Prob > old\_Prob$, then

   $old\_Prob \leftarrow new\_Prob$

   $ord \leftarrow ord + 1$

   $old\_\pi_i \leftarrow \pi_i\ \forall i, 1 \leq i \leq n$

   Discard $G$

   Goto Step 2

   Else goto Step 10

10. Output $old\_\pi_i\ \forall i, 1 \leq i \leq n$

    Output $Old\_Prob$

## 4  PRELIMINARY RESULTS

We used an implementation of the algorithm on a DEC Station 5000 to reconstruct the ALARM network (Figure 1) [Beinlich et. al., 89] by using 10,000 cases of a database generated by Herskovits ( [Herskovits, 91], [Cooper & Herskovits, 92] ). We used the $\chi^2$ test for the CI tests with a fixed $\alpha$ level of 0.1, and a bound of 15 on the maximum degree of the undirected graph generated in step 2. The algorithm recovered the network shown in Figure 2 using CI tests up to *only order 2*. Due to the bound, it did not generate a network for CI relations of order 0. Out of 46 edges, it recovered 45 edges (Figure 2).

---

[5] We use the notation $I(S_1, S_2, S_3)$ to represent the fact that $S_1$ and $S_3$ are independent conditional on $S_2$

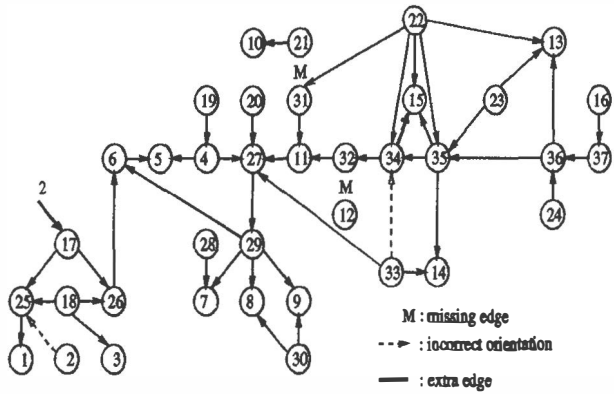

Figure 3: Constructed ALARM (Partial)

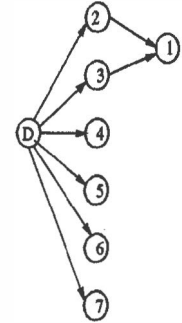

Figure 4: The LED Network

The only missing edge was the edge $12 \rightarrow 32$ (an edge which is not strongly supported by the data [Cooper & Herskovits, 92]). Two of the edges recovered were incorrectly oriented. However, the algorithm also recovered 14 extra edges. This is probably due to the incorrectly oriented edges, and to some extent, due to the greedy nature of K2. One of the incorrectly oriented edge was between the variables 34 and 33. As can be observed from Figure 2, 7 of the extra edges were between 33 and some other node. Moreover, an analysis of the order in which K2 selected the parents of node 37 showed that the 3 other extra edges incident on node 37 were recovered due to the greedy nature of K2 which, after picking node 16 as a parent of 37, picked up 33 because of the incorrect orientation, and then recovered the 3 edges of node 37 with 24, 23 and 22 once again due to its greedy search regimen. Similarly, the three extra edges involving node 2, 17 and 18 were recovered due to the fact that the edge between 2 and 25 was incorrectly oriented. The remaining extra edge was between nodes 15 and 34 which is recovered, once again, due to the greedy nature of K2. The total time taken was under 13 minutes.

[Cooper & Herskovits, 92] reported that K2, when given a total ordering consistent with the partial order of the nodes as specified by ALARM, recovered the complete network with the exception of one missing edge (between nodes 12 and 32) and one extra arc (from node 15 to 34). [Spirtes, 93] reported similar results with the PC algorithm. They applied the

PC algorithm [Spirtes & Glymour, 91] to the ALARM database split into two parts of 10000 cases each. The algorithm did not make any linearity assumption. In one case, the recovered network had no extra edge but had two missing edges while in the other case, the network had one extra edge and two missing edges.

To reduce the computational time, and to try to prevent the recovery of extra edges, we modified the algorithm by deleting step 7 of the algorithm. Instead of using a total order, K2 used a partial order defined on the nodes by the DAG constructed by step 6. The sets $\pi_i, 1 \leq i \leq n$, constructed by step 6 were given as input to K2 with the constraint that each node $i$ could have parents only from the set $\pi_i$. The network recovered by the algorithm after having used CI relations of *up to only order 2* is shown in Figure 3. It recovered 44 edges (the extra missing edge being $21 \rightarrow 31$); there were 2 extra edges (between 2 and 17, and between 34 and 15) while 2 edges were incorrectly oriented. However, the metric used by K2 ranked the earlier network structure (Figure 2) to be more probable. The time taken was reduced to under 7 minutes.

We also used the algorithm to reconstruct the faulty LED network (Figure 4) using a database of 199 cases ( [Fung & Crawford, 90] ). With an $\alpha$ value of 0.1, CB reconstructed the network (Figure 5) with 3 edges incorrectly oriented and one extra edge in less than 1 second using CI tests up to order 1. A subsequent analysis of the independence statements computed by CB found that the three incorrectly oriented edges were due to perceived independence of the pairs $(3,5)$, $(3,6)$, and $(4,5)$. While the underlying model did not support these independence statements, the data did. Step 3 oriented the edges according to the perceived independence. When we ran the modified version of CB using the partial order, the same network was recovered, except for the fact that there was no extra edge (Figure 5).

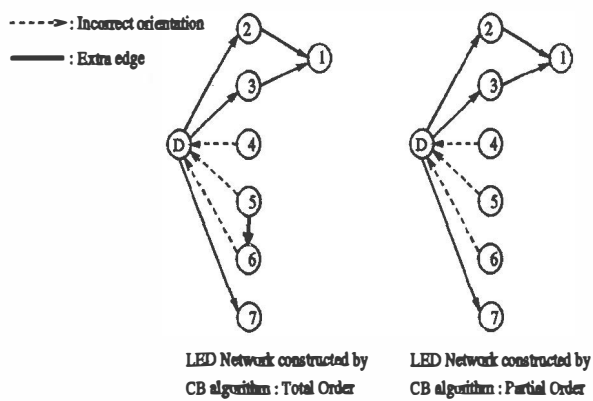

Figure 5: The Constructed LED Networks

# 5    SUMMARY AND OPEN PROBLEMS

In this paper, we have presented a method of recovering the structure of Bayesian belief networks from a database of cases by integrating CI test based methods and Bayesian methods.

Although these results are preliminary, they are quite encouraging because they show that the **CB** algorithm can recover a reasonably complex Bayesian network structure from data using **substantially low order CI relations**. Moreover, since it generates an ordering on the nodes from the database of cases only, without any outside information, it **eliminates the requirement for the user to provide an ordering on the variables**.

In the worst case, the CB algorithm is exponential in the number of variables, as explained below. Steps 1 (initialization) and 10 (output) of the algorithm are executed only once. The number of times that steps 2 through 9 of the CB algorithm are executed is bound by the sum of the largest two degrees in the undirected graph constructed at the end of step 2, by an argument almost identical to that of [Spirtes & Glymour, 91, page 68]. Each of steps 3 through 9 have only polynomial complexity in the number of variables, by arguments that are either simple or described in [Verma & Pearl, 92], [Cooper & Herskovits, 92]. In step 2, the number of independence tests carried out is exponential in the size of the order of the independence relations to be tested, which is bounded by the maximum of $|A_Gab|$. Note that the CB algorithm is polynomial for graphs for which $|A_Gab|$ is constant as the number of vertices increases, i.e. sparse graphs. *Our results indicate that the CB algorithm recovers Bayesian network structures in polynomial time in the number of domain variables, because the highest order of independence relations to be tested is very low.*

Although CB works well on the ALARM and LED networks and appears to be quite promising, a number of issues that could improve the performance of the algorithm need to be looked in further. We are already working on some of these issues.

Firstly, the CB algorithm has not yet been tested on large unknown databases. We are currently testing the CB algorithm on a number of databases that we have procured from the University of California (Irvine), Repository of Machine Learning databases. We also intend to test the algorithm on a large 147 variable medical database (cf. [Mechling & Valtorta, 93]), and see whether the recovered network is found plausible by medical experts.

Secondly, we have used a fixed $\alpha$ level for the $\chi^2$ test. This will almost certainly introduce dependencies that are purely the result of chance. It is possible to use the technique of Cross Validation for tuning this parameter. [Fung & Crawford, 90] discusses the tuning of the alpha level in performing belief-network learning.

Thirdly, the CB algorithm uses a greedy search mechanism (K2) to search for the set of parents of each node. This greedy search strategy does not ensure optimality even though the metric used by K2 is exact. Therefore, there is a need to explore other (less myopic) search methods like simulated annealing etc.

Also, since the quality of the recovered network structure is very sensitive to the ordering determined by phase I of the CB algorithm, efforts need to be made to find better and more efficient heuristics than the one presented in this paper that enable the selection of one orientation of an undirected edge over the other, since in general there will be a number of such undirected edges after steps 3 and 4 of the algorithm.

Moreover, most of the steps of the CB algorithm are inherently parallel. Hence, a huge reduction in the time required to recover the network structure can be possibly obtained by parallelizing the CB algorithm.

Finally, the CB algorithm uses a greedy strategy as a stopping criteria. It uses the probability of the entire network, as measured by the K2 metric, to decide when to stop; the algorithm stops when the value of the metric for the entire network is less than the value which had been computed for the network structure recovered in the previous iteration (i.e for a lower order of the CI tests). There is a need to look into alternative methods of terminating the algorithm.

### Acknowledgements

We are thankful to Prof. G. Cooper for providing the ALARM network database and to Dr. R. Fung for providing the LED network database. We are also grateful to the anonymous referees for their helpful comments and suggestions for improving the paper.

## References

[Beinlich et. al., 89] Beinlich, I.A., Suermondt, H.J., Chavez, R.M. and Cooper, G.F. "The ALARM monitoring system: A Case Study with Two Probabilistic Inference Techniques for Belief Networks", *Proceedings of the Second European Conference on Artificial Intelligence in Medicine*, 247-256, 1989, London, England. (as referenced in [Cooper & Herskovits, 92]).

[Cooper & Herskovits, 92] Cooper, G.F. and Herskovits, E. "A Bayesian Method for the Induction of Probabilistic Networks from Data", *Machine Learning*, 9, 309-347, 1992, Kluwer Academic Publishers.

[Fung & Crawford, 90] Fung, R.M. and Crawford, S.L. "Constructor: A System for the Induction of Probabilistic Models", *Proceedings of AAAI*, 762-769, 1990, Boston, MA: MIT Press

[Geiger and Heckerman, 91] Geiger, Dan and Heckerman, David. "Advances in Probabilistic Reasoning." *Uncertainty in Artificial Intelligence: Proceedings of the Seventh Conference*, San Mateo, CA: Morgan Kaufmann, 118–126, 1991.

[Glymour, et al., 1987] Glymour, C., Scheines, R., Spirtes, P., and Kelly, K. *"Discovering Causal Structure"*. San Diego, CA: Academic Press, 1987.

[Herskovits, 91] Herskovits, E., H. *"Computer-based probabilistic-network construction"*, Doctoral dissertation, Medical Information Sciences, Stanford University, Stanford, CA

[Lauritzen and Wermuth, 89a] Lauritzen, S.L. and N. Wermuth. "Graphical Models for Associations Between Variables, Some of Which Are Qualitative and Some Quantitative", *Annals of Statistics*, 17, 31–57, 1989.

[Lauritzen and Wermuth, 89b] Lauritzen, S.L. and N. Wermuth. "Graphical Models for Associations Between Variables, Some of Which Are Qualitative and Some Quantitative: Correction Note", *Annals of Statistics*, 17, 1916, 1989.

[Lauritzen, Thiesson, & Spiegelhalter, 93] Lauritzen S.L., B. Thiesson, and D. Spiegelhalter. "Diagnostic Systems Created by Model Selection Methods–A Case Study", *Preliminary Papers of the Fourth International Workshop on Artificial Intelligence and Statistics*, Ft. Lauderdale, FL, January 3-6, 93-105, 1993.

[Mechling & Valtorta, 93] Mechling, R. and Valtorta, M., "PaCCIN: A Parallel Constructor of Markov Networks", *Preliminary Papers of the Fourth International Workshop on Artificial Intelligence and Statistics*, Ft. Lauderdale, FL, January 3-6, 405-410, 1993.

[Pearl, 88] Pearl, Judea. *"Probabilistic Reasoning in Intelligent Systems"*, 1988, Morgan Kaufman, San Mateo.

[Pearl & Verma, 91] Pearl, Judea and Verma, Thomas. "A Theory of Inferred Causation", In Allen, J.A., Fikes, R., and Sandwell, E., editors, *Principles of Knowledge Representation and Reasoning: Proceedings of the Second International Conference*, 441-452, 1991, Morgan Kaufmann, San Mateo.

[Pearl & Wermuth, 93] Pearl, Judea and Nanny Wermuth. "When Can Association Graphs Admit a Causal Interpretation? (First Report)" *Preliminary Papers of the Fourth International Workshop on Artificial Intelligence and Statistics*, Ft. Lauderdale, FL, January 3-6, 141-150, 1993.

[Shachter, 91] Shachter, Ross D. "A Graph-Based Inference Method for Conditional Independence", *Uncertainty in Artificial Intelligence: Proceedings of the Seventh Conference*, San Mateo, CA: Morgan Kaufmann, 353–360, 1991.

[Spirtes, 93] Personal communication.

[Spirtes, Glymour & Scheines, 90] Spirtes, P., Glymour, C., and Scheines, R. "Causality from probability", In Tiles, J., McKee, G. and Dean, G., editors, *Evolving knowledge in the natural and behavioral sciences*, 181-199, 1990, London:Pitman.

[Spirtes & Glymour, 91] Spirtes, Peter and Glymour, Clark. "An Algorithm for Fast Recovery of Sparse Causal Graphs", *Social Science Computing Review*, 9:1, 62-72, 1991.

[Verma & Pearl, 92] Verma, Thomas and Pearl, Judea. "An Algorithm for Deciding if a Set of Observed Independencies Has a Causal Explanation", *Proceedings 8th Conference on Uncertainty in AI*, 323-330, 1992.